\documentclass[10pt,twocolumn,letterpaper]{article}

\usepackage{icb}
\usepackage{times}
\usepackage{epsfig}
\usepackage{graphicx}
\usepackage{amsmath}
\usepackage{amssymb}
\usepackage{multirow}
\usepackage[norule,symbol,perpage]{footmisc} 

\icbfinalcopy 

\ificbfinal\fi

\makeatletter
\def\ps@IEEEtitlepagestyle{
\def\@oddfoot{\mycopyrightnotice}
\def\@evenfoot{}
}
\def\mycopyrightnotice{
{\hfill \footnotesize 978-1-7281-3640-0/19/\$31.00 \copyright 2019 IEEE\hfill}
}
\makeatother

\begin{document}

\title{Attribute-Guided Deep Polarimetric Thermal-to-visible Face Recognition}

\author{Seyed Mehdi Iranmanesh, Nasser M. Nasrabadi\\
West Virginia University\\
{\tt\small \{seiranmanesh\}@mix.wvu.edu, \{nasser.nasrabadi\}@mail.wvu.edu}}
\maketitle
\thispagestyle{empty}

\begin{abstract}
	In this paper, we present an attribute-guided deep coupled learning framework to address the problem of matching polarimetric thermal face photos against a gallery of visible faces. The coupled framework contains two sub-networks, one dedicated to the visible spectrum and the second sub-network dedicated to the polarimetric thermal spectrum. Each sub-network is made of a generative adversarial network (GAN) architecture.  We propose a novel Attribute-Guided Coupled Generative Adversarial Network (AGC-GAN) architecture which utilizes facial attributes to improve the thermal-to-visible face recognition performance. The proposed AGC-GAN exploits the facial attributes and leverages multiple loss functions in order to learn rich discriminative features in a common embedding subspace. To achieve a realistic photo reconstruction while preserving the discriminative information, we also add a perceptual loss term to the coupling loss function. An ablation study is performed to show the effectiveness of different loss functions for optimizing the proposed method. Moreover, the superiority of the model compared to the state-of-the-art models is demonstrated using polarimetric dataset. 
\end{abstract}
\let\thefootnote\relax\footnotetext{\mycopyrightnotice} 
\section{Introduction}

In recent years, there has been significant amount of research in Heterogeneous Face Recognition (HFR)~\cite{3ICB}. The main issue in HFR is to match the visible face image to a face image that has been captured in another domain such as in the infrared spectrum~\cite{3ICB}, or the polarimetric thermal~\cite{5ICB} domain. Infrared images are categorized into two major groups of reflection and emission. The reflection category, which contains near infrared (NIR) and shortwave infrared (SWIR) bands, is more informative about the facial details and it is very similar to the visible imagery. Due to this reflective phenomenology of the NIR and SWIR, there has been a significant performance on NIR-to-visible face recognition accuracy~\cite{7ICB} and to some extent for SWIR-to-visible face recognition accuracy~\cite{8ICB}. 

\begin{figure}
	\begin{center}
		\includegraphics[width=0.95\linewidth]{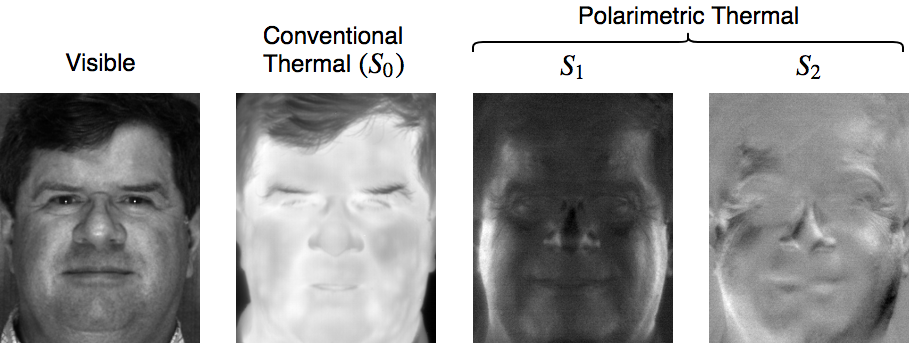}
		
	\end{center}
	\caption{Visible spectrum and its corresponding conventional thermal ($S_0$), and polarimetric state information ($S_1$ and $S_2$) of a thermal image of a subject.}
	\label{fig1_ICB}
\end{figure}

The emission category contains the midwave infrared (MWIR) and longwave infrared (LWIR) bands and it is less informative~\cite{3ICB} compared with the reflection category. Due to the significant phenomenological differences between the distribution of thermal and visible imagery, matching a thermal face against a gallery of visible faces becomes a challenging task. However, thermal-to-visible recognition is highly demanding because in the thermal data no active illumination is needed at night-time or low-light environments since the thermal imagery is based on the emission and originates from the underlying skin and depends on the individual's physiology. Many recent approaches on cross-modal problem have mainly focused on closing the gap between the two different modalities, but they have not explored the soft biometrics i.e., facial attributes. 
\begin{figure*}
	\begin{center}
		\includegraphics[width=0.8\linewidth]{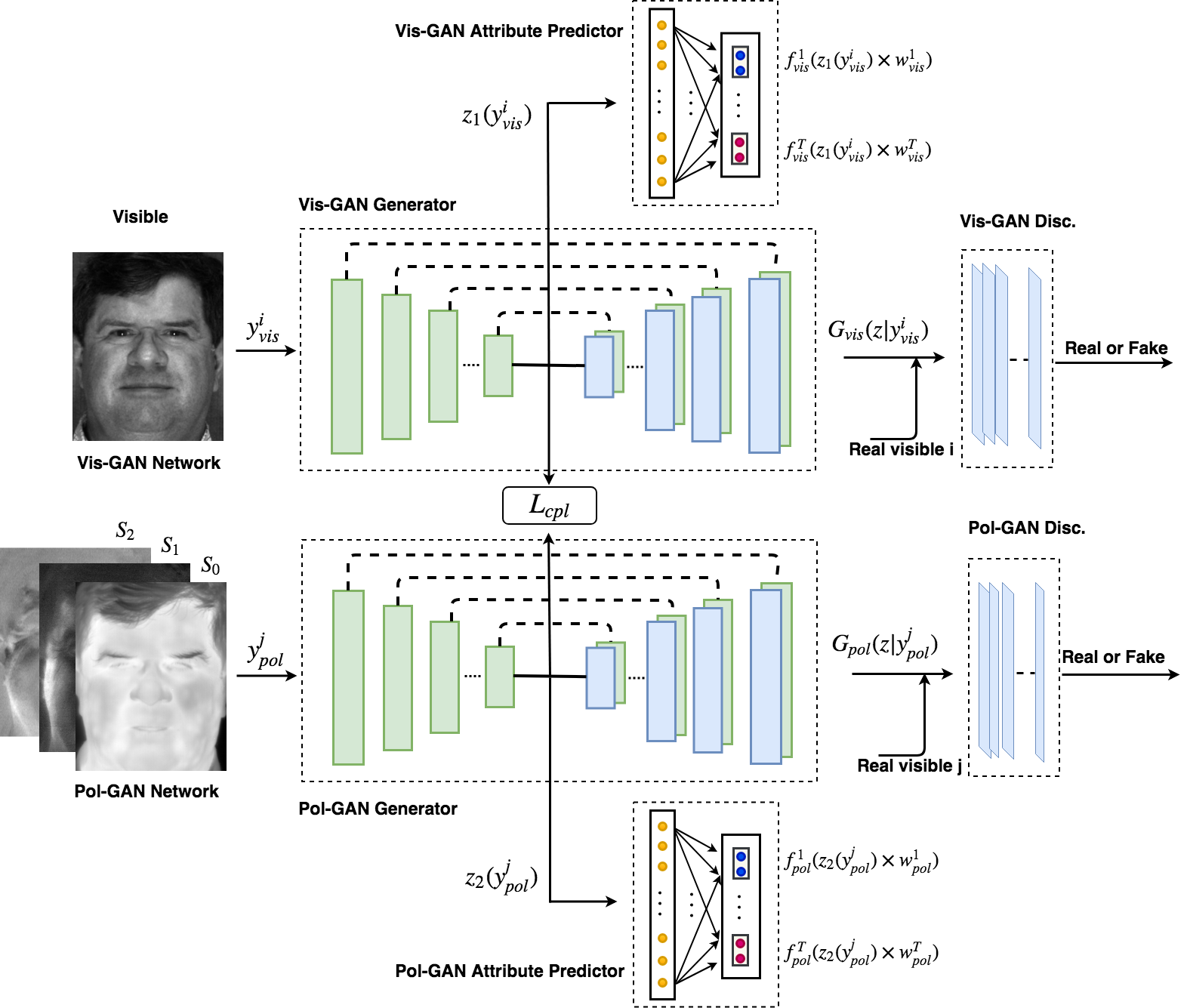}
		
	\end{center}
	\caption{Proposed network using two GAN-based sub-networks (Vis-GAN and Pol-GAN) coupled by a contrastive loss function. Input to the Pol-GAN is the polarimetric data ($S_0,S_1,S_2$). The facial attributes are predicted from both sub-networks (Vis-GAN, Pol-GAN) in a multi-tasking paradigm.}
	\label{fig2_ICB}
\end{figure*}

Recently, via an emerging technology~\cite{5ICB}, the polarization state information of thermal emission has been exploited to provide additional geometrical and textural details, especially around the nose and the mouth, which complements the textural details of the conventional intensity-based thermal images. This additional information which is not available in the conventional intensity-based thermal imaging~\cite{5ICB}, is utilized in recent algorithms to enhance the cross thermal-to-visible face recognition~\cite{10ICB}. Fig.~\ref{fig1_ICB} shows a visible image and its corresponding conventional thermal ($S_0$) and polarimetric state information ($S_1$ and $S_2$) images. 

There are some informative traits, such as age, gender, ethnicity, race, and hair color, which are not distinctive enough for the sake of recognition, but still can act as complementary information to other primary information, such as face and fingerprint. These traits, which are known as soft biometrics, can improve recognition algorithms while they are much cheaper and faster to acquire. They can be directly used in a unimodal system for some applications. Soft biometric traits have been utilized jointly with hard biometrics (face photo) for different tasks such as person identification or face recognition~\cite{12ICB}. However, the soft biometric traits are considered to be available both during the training and testing phases. Our approach looks at this problem in a different way. We consider the case when soft biometric information does not exist during the testing phase, and our method can predict them directly in a \textit{multi-tasking} paradigm.


Multi-task learning (MTL) and deep learning techniques has been vastly applied in computer vision and biometrics problems~\cite{mehrizi2019predicting,mehrizi2018toward,mehrizi2018computer}. MTL basically attempts to solve correlated tasks concurrently with the help of knowledge sharing between tasks. In~\cite{17ICB} MTL is employed to predict attributes such as age, gender, race, etc.  Face photo can be viewed as having some positive or negative hidden relationship with some of its soft facial biometric traits.

In this paper, we propose an Attribute-Guided Coupled Generative Adversarial Network (AGC-GAN), which considers the Convolutional Neural Network (CNN) weight sharing followed by the dedicated weights which are responsible for learning the representative features for each specific face attribute. The network also tries to find the common embedding space between the polarimetric thermal and visible faces utilizing a coupling structure and an adversarial training. Optimizing the coupled network with the help of the facial attributes leads to a  more discriminative embedding subspace and can be utilized to enhance the performance of the main task which is the heterogeneous face recognition.

To summarize, the following are our main contributions: 

\begin{description}
	\item[$\bullet$] A novel polarimetric thermal-to-visible face recognition system is proposed based on an attribute-guided coupled GAN to synthesize visible faces from the polarimetric thermal images. 
	
	\item[$\bullet$] A multi-tasking framework is proposed to predict facial attributes from the polarimetric thermal faces. To the best of our knowledge, no such demonstration has been proposed in the literature. 
	
	\item {$\bullet$} Extensive experiments are conducted on ARL polarimetric facial database~\cite{5ICB} and the proposed method is compared to recent state-of-the-art methods. 
\end{description}

\section{Preliminaries}

\begin{figure}
	\begin{center}
		\includegraphics[width=0.8\linewidth]{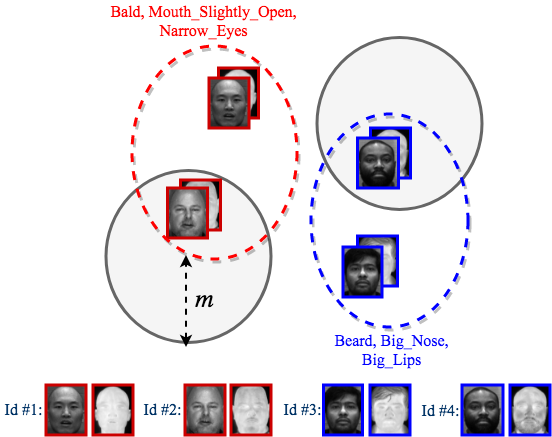}
		
	\end{center}
	\caption{Visualization of the common latent subspace by leveraging facial attributes prediction loss function. Solid circles represent the contrastive margin and the dashed circles depict the attributes classification. For the sake of clarity the contrastive margin is depicted for two $Ids$ out of four $Ids$.}
	\label{fig3:latent}
\end{figure}




\subsection {Conditional Generative Adversarial Networks}

The generative adversarial network consists of two networks, namely a generator and discriminator which compete with each other in a minimax game~\cite{20ICB}. Conditional adversarial networks is an extension of generative adversarial networks in which both the generator and discriminator are conditioned on some auxiliary information $y$. The extra information $y$ can be any kind of information such as class label or data from other modalities. The objective of the conditional GAN is the same as the classical GAN as follows:
\vspace*{-2mm}
\begin{align}\label{ConditionGAN}
	&\underset{G}{\operatorname{min}}\, \underset{D}{\operatorname{max}} E_{x\sim P_{data(x)}}[log D(x|y)]+\\ &E_z\sim P_z [log(1-D(G(z|y)))] \nonumber .
\end{align}

\section {Proposed Method}

The proposed AGC-GAN is illustrated in Fig.~\ref{fig2_ICB}. The proposed approach consists of two coupled generators and two discriminators. Each generator is also responsible to predict facial attributes in a multi-tasking paradigm. In the following subsections we explain these modules in detail.

\subsection{Deep Coupled Framework}

The final objective of the proposed model is the identification of the polarimetric test faces against a gallery of visible faces which we do not have access to them during the training phase. For this reason, we couple two U-net networks~\cite{21ICB} one is dedicated to the visible spectrum (Vis-GAN) and the other network is dedicated to the polarimetric spectrum (Pol-GAN). Each network performs a non-linear transformation of the input space. The final objective of our proposed AGC-GAN is to find the global deep latent features representing the relationship between polarimetric face images and their corresponding visible ones. In order to find a common latent embedding subspace between these two different domains we couple two networks (Vis-GAN and Pol-GAN) via a contrastive loss function~\cite{22ICB}. This loss function ($\ell_{cont}$) pulls the genuine pairs (i.e., a visible face image with its own corresponding polarimetric face images) towards each other in a common latent feature subspace and push the impostor pairs (i.e., a visible face image of a subject with another subject's polarimetric face images) apart from each other (see Fig.~\ref{fig2_ICB}). 
Similar to~\cite{22ICB}, our contrastive loss is of the form:
\begin{align}\label{contrastiveloss}
	\ell_{cont}&(z_1(y^i_{vis}),z_2(y^j_{pol}),y_{cont})=\\ \nonumber &(1-y_{cont})L_{gen}(D(z_1(y^i_{vis}),z_2(y^j_{pol}))+\\ \nonumber & y_{cont}L_{imp}(D(z_1(y^i_{vis}),z_2(y^j_{pol})) \; ,
\end{align}
where $y^i_{vis}$ is the input for the Vis-GAN (i.e., visible face image), and $y^j_{pol}$ is the input for the pol-GAN (i.e., polarimetric face images). $y_{cont}$ is a binary label, $L_{gen}$ and $L_{imp}$ represent the partial loss functions for the genuine and impostor pairs, respectively, and $D(z_1(y^i_ {vis}),z_2(y^j_{pol}))$ indicates the Euclidean distance between the embedded data in the common latent feature subspace. $z_1(.)$ and $z_2(.)$ are the deep CNN-based embedding functions, which transform $y^i_{vis}$ and $ y^j_{pol}$ into a common latent embedding subspace. The binary label, $y_{cont}$, is assigned a value of 0 when both modalities, i.e., visible and polarimetric, form a genuine pair, or, equivalently, the inputs are from the same class ($cl^i =cl^j$). On the contrary, when the inputs are from different classes, which means they form an impostor pair, $y_{cont}$ is equal to 1. In addition, $L_{gen}$ and $L_{imp}$ are defined as follows: 
\begin{equation}
	\begin{split}
		& L_{gen}(D(z_1(y^i_{vis}),z_2(y^j_{pol})))= \\&\dfrac{1}{2}  ||z_1(y^i_{vis})-z_2(y^j_{pol})||^2_2
		\quad \quad  \text{for} \quad\quad cl^i=cl^j  \;,
		\label{eq-sda-separation}
	\end{split}
\end{equation}
\noindent and
\begin{align}
	& L_{imp}(D(z_1(y^i_{vis}), z_2(y^j_{pol})))=\\ \nonumber
	& \dfrac{1}{2} \:  max(0,m-||z_1(y^i_{vis})-z_2(y^j_{pol})||^2_2)\;\;  \text{for} \quad cl^i \neq cl^j,
	\label{eq-sda-separation} 
\end{align}

\noindent where $m$ is the contrastive margin. Therefore, the coupling loss function can be written as:
\begin{equation}
	\begin{split}
		L_{cpl} &= 1/N^2 \displaystyle\sum_{i=1}^{N}\displaystyle\sum_{j=1}^{N} \ell_{cont}(z_1(y^i_{vis}), z_2(y^j_{pol}), y_{cont}),
		\label{eq-5-cont}
	\end{split}
\end{equation}

\noindent where $N$ is the number of training samples. It should be noted that the contrastive loss function~(\ref{eq-5-cont}) considers the subjects' labels implicitly. Therefore, it has the ability to find a discriminative embedding space by employing the data labels in contrast to some other metrics such as the Euclidean distance. This discriminative embedding subspace would be useful in identifying a polarimetric probe photo against a gallery of visible face photos.

\subsection{Multi-Attribute Prediction and Identification Task:}

The objective of this model is to predict a set of attributes using a visible or polarimetric face images. Therefore, in this architecture a visible face image (polarimetric face images) is presented to the network as an input and a set of attributes are predicted. Suppose the input is a visible image $y^i_{vis} \in Y$, and its class label is $cl^i \in CL$ for $i=1,\dots,N$ where $N$ is the number of the training samples. Let the soft biometric traits, contain $T$ different facial attributes or binary class labels. Therefore, in this framework we denote them as $ cl^{t}$ for $t=1,\dots,T$ . Learning multiple CNNs separately is not optimal since different tasks may have some hidden relationships with each other and may share some common features. This is supported by~\cite{23ICB} where they train a CNN features for the face recognition task and they use it directly for the facial attribute prediction. Therefore, our network shares a big portion of its parameters among different attribute prediction tasks in order to enhance the performance of the recognition task. Thus, the loss function is as follows: 

\begin{equation}
	\begin{split}
		L_{avis} &= 1/N \displaystyle\sum_{i=1}^{N} \displaystyle\sum_{t=1}^{T}\ell(f^t_{vis}(z_1(y_{vis}^i)\times w_{vis}^{t}), cl^{i,t})\; ,
	\end{split}
	\label{label-2}
\end{equation}

\noindent where $\ell$ is a proper loss function (e.g., cross entropy) and $f^t_{vis}(.)$ is a binary classifier for the attribute $t$ operated on the bottleneck of Vis-GAN (see Fig.~\ref{fig2_ICB}). $w_{vis}^{t}$ represents the remaining parameters which are learned separately for each facial attribute task.

The same procedure is performed in the other network (Pol-GAN) with polarimetric thermal images as input. The Pol-GAN network is also responsible to estimate a set of facial attributes. Therefore, the loss function is: 

\begin{equation}
	\begin{split}
		L_{apol} =  &1/N \displaystyle\sum_{j=1}^{N} \displaystyle\sum_{t=1}^{T}\ell (f^t_{pol}(z_2(y^j_{pol})\times w_{pol}^t), cl^{j,t})\; ,
	\end{split}
	\label{label-4}
\end{equation}

\noindent where $\ell$ is the cross entropy loss function and $f^t_{pol}(.)$ is a binary classifier for the attribute $t$ operated on the bottleneck of Pol-GAN (see Fig.~\ref{fig2_ICB}). $w_{pol}^t$ represents the remaining features which are assigned separately for each facial attribute prediction task. The total attribute prediction loss function is: 
\vspace*{-2mm}
\begin{equation}
	\begin{split}
		L_{a} = L_{avis} + L_{apol}.\; 
	\end{split}
	\label{label-5}
\end{equation}

\vspace*{-2mm}

\subsection{Generative Adversarial Loss}

Let $G_{vis}$ and $G_{pol}$ denote the generators that synthesize the corresponding visible image from the visible and its polarimetric images, respectively. To synthesize the output and to make sure that the synthesized images generated by the two generators are indistinguishable from the corresponding ground truth visible image, we utilized the GAN loss function in~\cite{20ICB}. As it is shown in Fig.~\ref{fig2_ICB}, the first generator $G_{vis}$ is responsible to generate a visible image when the network is conditioned on a visible image. On the other hand, the second generator $G_{pol}$ tries to generate the visible image from the polarimetric images which is a more challenging task compared to the first generator. Therefore, the total loss for the coupled GAN is as follows: 

\vspace*{-4mm}
\begin{align}\label{total_GAN_1}
	&L_{GAN} = L_{vis} + L_{pol}  ,
\end{align}

\noindent where the GAN loss function for the Vis-GAN sub-network is given as:
\vspace*{-2mm}
\begin{align}\label{Vis-GAN}
	&L_{vis} = \underset{G_{vis}}{\operatorname{min}}\, \underset{D_{vis}}{\operatorname{max}} E_{x^i\sim P_{vis(x)}}[log D(x^i|y^i_{vis})]+\\ &E_z\sim P_z [log(1-D(G(z|y^i_{vis})))] \nonumber ,
\end{align}

\noindent where $y^i_{vis}$ is the visible image used as condition for the Vis-GAN and $x^i$ is the real visible data. It should be noted that for the Vis-GAN the real data $x^i$ and the condition $y^i_{vis}$ are the same. Similarly the loss for the Pol-GAN is given as: 
\vspace*{-2mm}
\begin{align}\label{Pol-GAN}
	&L_{pol} = \underset{G_{pol}}{\operatorname{min}}\, \underset{D_{pol}}{\operatorname{max}} E_{x^j\sim P_{vis(x)}}[log D(x^j|y^j_{pol})]+\\ &E_z\sim P_z [log(1-D(G(z|y^j_{pol})))] \nonumber ,
\end{align}

\noindent where $y^j_{pol}$ is the polarimetric images used as condition for the Pol-GAN and $x^j$ is the real visible data. It should be noted that $x^i$ is the same as $x^j$ if they refer to the same person ($cl^i=cl^j$) and otherwise they are not the same.

\subsection{Overall Loss Function}\label{attributess} 

The proposed approach contains the following loss functions: the Euclidean $L_{E_{vis}}$ and $L_{E_{pol}}$ losses which are enforced on the recovered visible images from the Vis-GAN and Pol-GAN sub-networks, respectively, are defined as follows:
\vspace*{-2mm}
\begin{align}\label{total_GAN}
	&L_{E_{vis}}= ||G_{vis}(z|y^i_{vis})-x^i||^2_2 , \\&
	L_{E_{pol}}= ||G_{pol}(z|y^j_{pol})-x^j||^2_2 , \\&
	L_{E}= L_{E_{vis}} + L_{E_{pol}}.
\end{align}

The $L_{GAN}$~(\ref{total_GAN_1}) loss is also added to generate sharper images. In addition, based on the success of the perceptual loss in low-level vision tasks~\cite{24ICB}, a perceptual loss is added to the Pol-GAN sub-network to generate a more realistic face photos as follows: 
\vspace*{-2mm}
\begin{align}\label{total_GAN}
	& L_{P_{pol}} = \tfrac{1}{C_pW_pH_p}\displaystyle\sum_{c=1}^{C_p}\displaystyle\sum_{w=1}^{W_p}\displaystyle\sum_{h=1}^{H_p}\\&||V(G_{pol}(z|y^j_{pol}))^{c,w,h}-V(x^j)^{c,w,h}|| \nonumber,
\end{align}

\noindent where $x^j$ is the ground truth visible image, $G_{pol}(z|y^j_{pol})$ is the output of Pol-GAN generator. V(.) represents a non-linear CNN transformation and $C_p, W_p, H_p$ are the dimension of a particular layer in $V$. It should be noted that the perceptual loss is just used in the Pol-GAN sub-network. 
Similarly, we utilized a perceptual attribute loss which measures the difference between the facial attributes of the synthesized images and the real image. To extract attributes from a given visible face, we fine-tune the pre-trained VGG-Face~\cite{25ICB} on ten annotated facial attributes as tabulated in Table~\ref{tableprediction}. This network (attribute predictor) is trained separately from AGC-GAN. Afterward, this attribute predictor is utilized for perceptual attribute loss on Vis-GAN and Pol-GAN as follows:
\vspace*{-1mm}
\begin{align}\label{preceptual_attribute}
	& L_{pavis} = ||A(G_{vis}(z|y_{vis}^i))-A(x^i)||_2^2, \\ &
	L_{papol} = ||A(G_{pol}(z|y_{pol}^j))-A(x^j)||_2^2,  \\ &
	L_{pa} = L_{pavis} + L_{papol} ,
\end{align}

\noindent where $A$ is the fine-tuned VGG-Face attribute predictor network. $L_{pa}$ is the total perceptual attribute loss function which composed of the perceptual attribute losses for the Vis-GAN ($L_{pavis}$) and Pol-GAN ($L_{papol}$) sub-networks. 

Finally, the coupling loss function~(\ref{eq-5-cont}) is added to train both networks Vis-GAN and Pol-GAN jointly to make the embedding subspace of the mentioned networks as close as possible and to preserve a more discriminative and distinguishable shared space. Therefore, the total loss function is as follows: 
\vspace*{-2mm}
\begin{align}\label{total_GAN_t}
	& L_{T} = L_{cpl}+ \lambda_{1} L_{E}+ \lambda_2 L_{GAN} \\ &\lambda_3 L_{a} + \lambda_4 L_{Ppol} + \lambda_5 L_{pa}\nonumber,
\end{align}

\noindent where $\lambda_1$, $\lambda_2$, $\lambda_3$, $\lambda_4$, and $\lambda_5$ are the hyper-parameters which weight different loss terms in the total loss function.

\begin{figure}
	\begin{center}
		\includegraphics[width=1.1\linewidth]{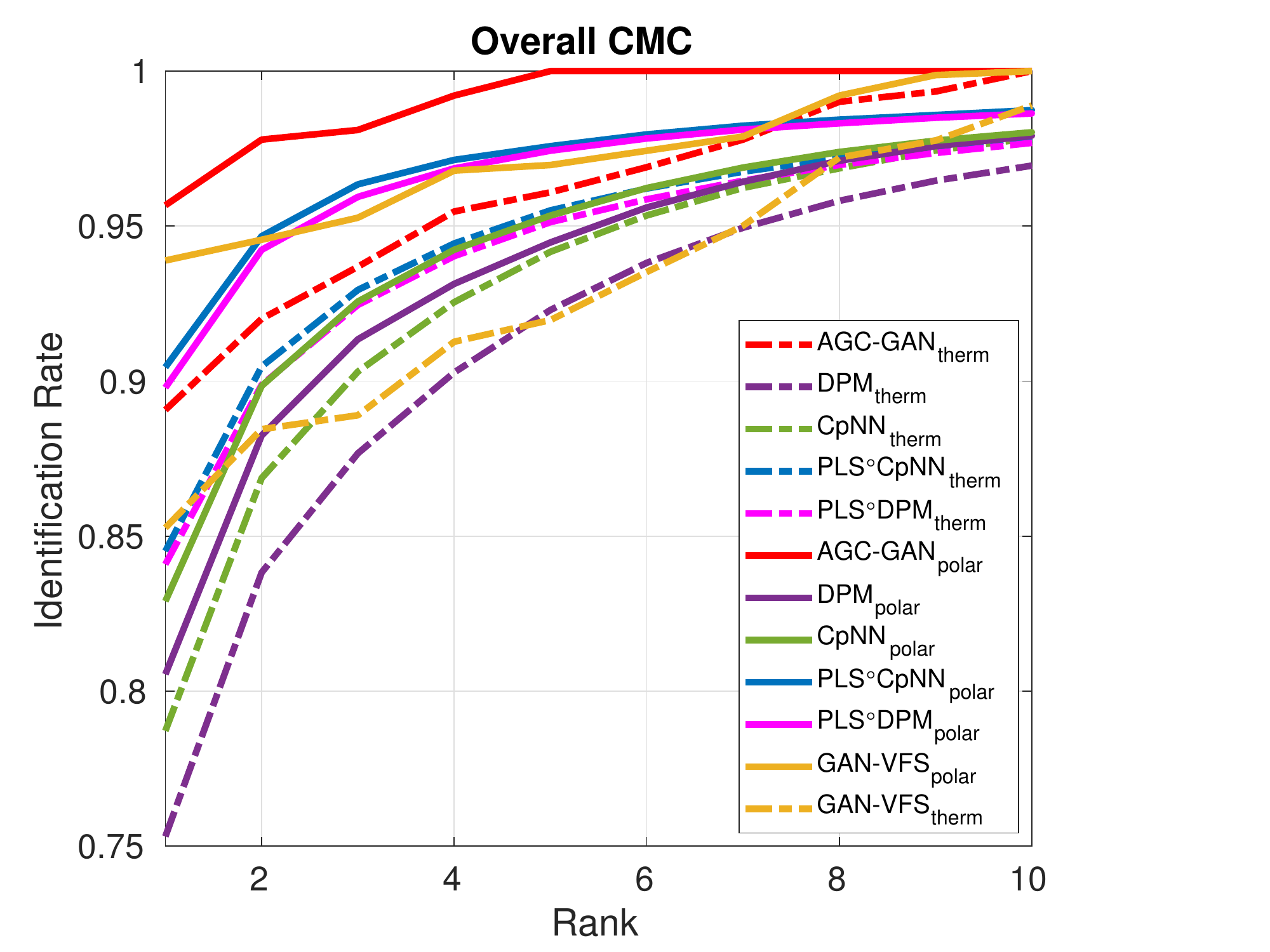}
		
	\end{center}
	\caption{Overall CMC curves from testing AGC-GAN versus other baselines using polarimetric and thermal probe samples.}
	\label{fig:cmc1}
\end{figure}

\subsection{Testing Phase}

During the testing phase, only the Pol-GAN is used. For a given test probe $y_{pol}^t$, Pol-GAN sub-network of the proposed AGC-GAN is employed to synthesize the visible image $G_{pol}(z|y^t_{pol})= \hat{x}^t_{vis}$. Eventually, the identification of face recognition is done, by calculating the minimum Euclidean distance between the synthesized image from the polarimetric prob against the visible gallery images as follows:     

\begin{equation}
	\begin{split}
		x^{t^*}_{vis} = \underset{x^t_ {vis}}{\operatorname{argmin}}\quad ||x^t_{vis},\hat{x}^t_{vis}||  \;,
		\label{eq-6}
	\end{split}
\end{equation}

\noindent where $\hat{x}^t_{vis}$ is the synthesized probe face image and $x^{t^*}_{vis}$ is the selected matching visible face image within the gallery of face images. In addition, the Pol-GAN sub-network can be employed to predict the facial attributes from the polarimetric face probe. The predicted facial attributes can also be used to narrow the search or identify a person of interest in a visible gallery of faces.     

\subsection{Implementation Details}

A U-net structure~\cite{21ICB} is employed as the network for the generator since it is able to address the vanishing gradient problem as well as capturing large receptive field. Also, a patch-based discriminator~\cite{26ICB} is used in the proposed method and it is trained iteratively with the generator. The entire network is trained in Pytorch. For the sake of training AGC-GAN, the hyper-parameters for all the loss functions are set to one except for the perceptual loss $L_{p_{pol}}$ and perceptual attribute loss $L_{pa}$  which is set to 0.5.  For training we used Adam optimizer~\cite{27ICB} with the first-order momentum of 0.5, the learning rate of 0.0002, and batch size of 4. For the generator the ReLU activation, and for the discriminator the Leaky ReLU activation with the slope of 0.2 is considered. The perceptual loss is assessed on the relu3-1 layer of the pre-trained VGG model~\cite{29ICB}. In order to fine-tune the attribute predictor network utilized for perceptual attribute loss, we manually annotated images with the attributes tabulated in Table~\ref{tableprediction}.

\section {Experiments}
We evaluate the proposed face recognition method by comparing with several recent works~\cite{30ICB,31ICB,32ICB,5ICB,34ICB,35ICB} on the ARL Multi-modal Face database~\cite{5ICB}.

\noindent{\bf Polarimetric Thermal Face} dataset~\cite{5ICB} comprises polarimetric thermal face images and their corresponding visible spectrum faces related to 60 subjects. Data was collected at three different distances: Range 1 (2.5 m), Range 2 (5 m), and Range 3 (7.5 m). At each range two different conditions, including baseline and expression are considered. 
\begin{table*}
\begin{center}
	\caption{Rank-1 identification rate for cross-spectrum face recognition using polarimetric thermal and conventional thermal ($S_0$) probe imagery.}
	\label{table:table1}
	\scalebox{0.8}{
		\begin{tabular}{|c|c|c|c|c|c|c|c|c|}
			\hline
			Scenario & \multicolumn{8}{c|}{Rank-1 Identification Rate} \\ \cline{2-9}
			\multirow{3}{*}{} & Probe  & PLS  & DPM  & CpNN & PLS$\circ$DPM & PLS$\circ$CpNN & GAN-VFS & AGC-GAN\\ \cline{2-9}  \hline  \hline
			Overall & Polar  &  0.5867 & 0.8054  & 0.8290 & 0.8979 &  0.9045& 0.9382 & \textbf{0.9654} \\ \cline{2-9} 
			& Therm  & 0.5305  & 0.7531  & 0.7872 & 0.8409 &  0.8452 & 0.8561 & \textbf{0.8925} \\ \hline
			\multirow{2}{*}{}Expressions & Polar  & 0.5658  &   0.8324& 0.8597  & 0.9565 & 0.9559  & 0.9473 & \textbf{0.9733} \\ \cline{2-9} 
			& Therm  & 0.6276  & 0.7887  & 0.8213 & 0.8898 &  0.8907 & 0.8934& \textbf{0.9217} \\ \hline
			\multirow{2}{*}{} Range 1 Baseline & Polar  & 0.7410  & 0.9092   & 0.9207 & 0.9646 & 0.9646 & 0.9653 & \textbf{0.9883} \\ \cline{2-9} 
			& Therm  & 0.6211  & 0.8778   & 0.9102  & 0.9417 & 0.9388  & 0.9412 & \textbf{0.9659} \\ \hline
			\multirow{2}{*}{} Range 2 Baseline & Polar  & 0.5570  & 0.8229  & 0.8489 & 0.9105 & 0.9187 & 0.9263 & \textbf{0.9643}  \\ \cline{2-9} 
			& Therm  & 0.5197  & 0.7532  & 0.7904 & 0.8578 &  0.8586 & 0.8701 & \textbf{0.9178} \\ \hline
			\multirow{2}{*}{} Range 3 Baseline & Polar  & 0.3396  & 0.6033  & 0.6253 & 0.6445 & 0.6739 & 0.8491 & \textbf{0.9068}  \\ \cline{2-9} 
			& Therm  & 0.3448  & 0.5219  & 0.5588 & 0.5768 & 0.6014  & 0.7559 & \textbf{0.8124}   \\ \hline
		\end{tabular}}
	\end{center}
\end{table*}

To increase the correlation between the two modalities of visible and thermal, each modality was preprocessed. We applied a band-pass filter so called difference of Gaussians (DoG), to emphasize the edges in addition to removing high and low frequency noise. 

We pass $S_0$, $S_1$, and $S_2$ to the Pol-GAN's three channels as the input as shown in Fig.~\ref{fig2_ICB}. To train the network, the genuine and impostor pairs are constructed. The genuine pair is constructed for the same subject photos from the two different modalities. For the impostor pair, a different subject is selected for each modality. In general, the number of the generated impostor pairs are significantly larger than the genuine pairs. For the sake of balancing the training set, we consider the same number of genuine and impostor pair. After training the model, during the testing phase, only the polarimetric sub-network is used for the evaluation. For a given probe, the Pol-GAN sub-network is used to synthesize the visible image. Afterwards, the Euclidean distance is used to match the synthesize image to its closest image from the gallery. The ratio of the number of correctly classified subjects and the entire number of subjects is computed as the identification rate. 

In each experiment the dataset is partitioned to the training and testing sets randomly. The same set of training and testing are used to evaluate PLS~\cite{30ICB}, DPM~\cite{31ICB}, CpNN~\cite{32ICB}, PLS$\circ$DPM~\cite{5ICB}, PLS$\circ$CpNN~\cite{34ICB}, GAN-VFS~\cite{35ICB}, and the proposed AGC-GAN network. Fig.~\ref{fig:cmc1} shows the overall cumulative matching characteristics (CMC) curves for our proposed method and the other state-of-the-art methods over all the three different ranges as well as the expressions data at Range 1. For the sake of comparison, in addition to the polarimetric thermal-to-visible face recognition performance, Fig.~\ref{fig:cmc1} also shows the results for the conventional thermal-to-visible face recognition for some of the methods, namely PLS$\circ$DPM, PLS$\circ$CpNN, CpNN, and AGC-GAN. In the conventional thermal-to-visible face recognition, all the mentioned methods exactly follow the same procedure as before, with only using $S_0$ modality. Fig.~\ref{fig:cmc1} illustrates that exploiting the polarization information of the thermal spectrum enhances cross-spectrum face recognition performance compared to the conventional one. Fig.~\ref{fig:cmc1} also shows the superior performance of our approach compared to the state-of-the-art methods. In addition, our method could achieve prefect accuracy (100$\%$) at Rank-4 and above. 
\begin{figure}[t]
	\begin{center}
		\includegraphics[width=0.9\linewidth]{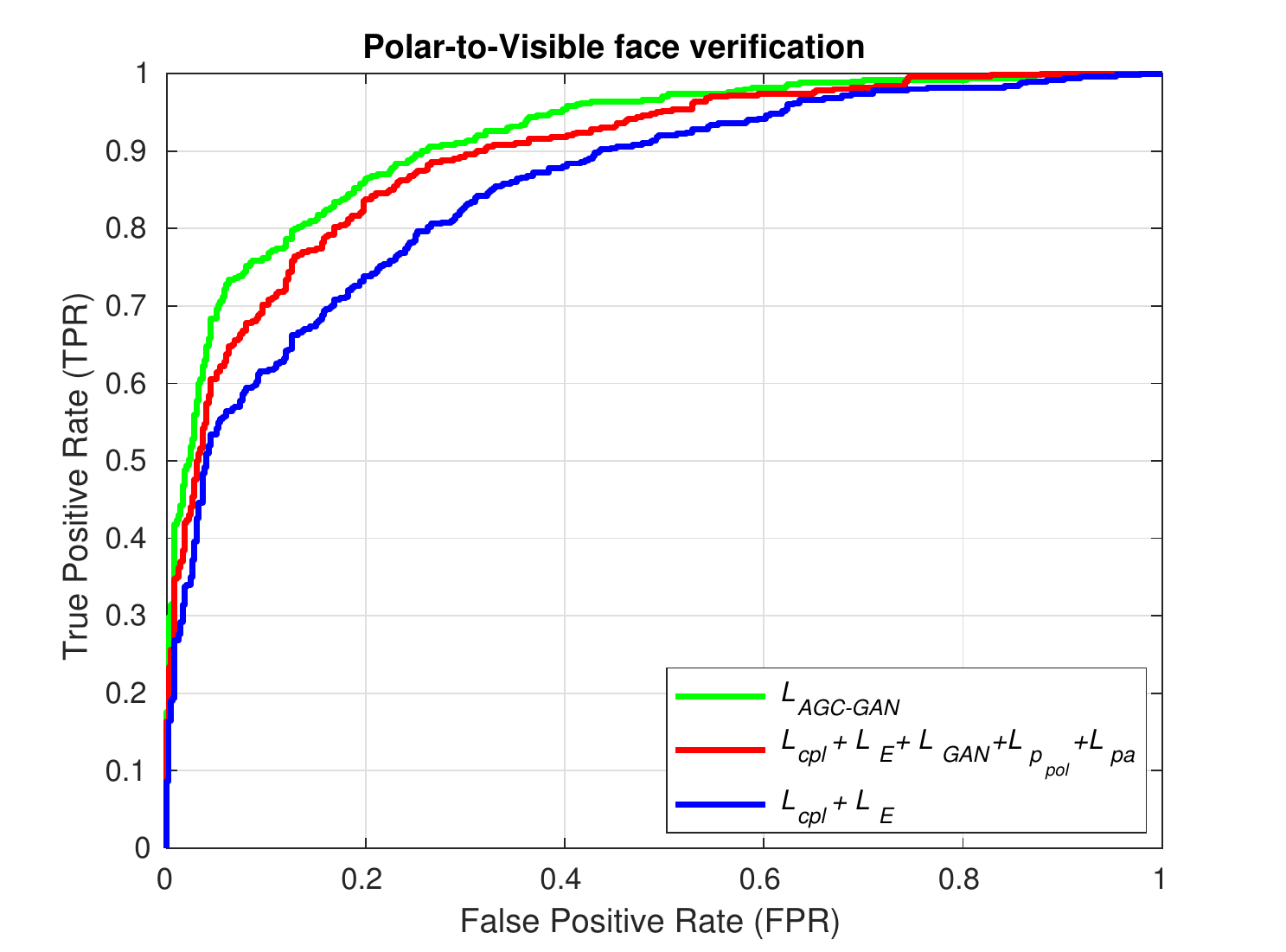}
		
	\end{center}
	\caption{The ROC curves corresponding to the ablation study.}
	\label{fig:ablation_2}
\end{figure}

Table~\ref{table:table1} tabulates the Rank-1 identification rates for five different scenarios: overall (which corresponds to Fig.~\ref{fig:cmc1}), Range 1 expressions, Range 1 baseline, Range 2 baseline, and Range 3 baseline. In our proposed approach, exploiting polarization information enhance the Rank-1 identification rate by 2.24\%, 5.16\%, 4.65\%, and 9.44\% for Range 1 baseline, Range 1 expression, Range 2 baseline, and Range 3 baseline compared to the conventional thermal-to-visible face recognition. This table reveals that using deep coupled generative adversarial network technique with the contrastive loss function as well as utilizing facial attributes to transform different modalities into a distinctive common embedding subspace is superior to the other embedding techniques such as PLS$\circ$CpNN. It also shows the effectiveness of our method in exploiting polarization information to improve the cross-spectrum face recognition problem.    

\section {Ablation Study}

In order to illustrate the effect of adding different loss functions and their improvement in our proposed framework, we perform a study with the following evaluations using the polarimetic dataset: 1) Polar-to-visible using the coupled framework with using only $L_{cpl}+ L_{E}$ losses, 2) Polar-to-visible using the proposed framework with $L_{cpl}+ L_{E}+ L_{GAN}+L_{p_{pol}}+L_{pa}$ loss functions, and 3) Polar-to-visible with all the loss functions in the proposed framework~(\ref{total_GAN_t}).

We plot the receiver operation characteristic (ROC) curves corresponding to the mentioned three different settings of the framework in the task of face verification. As it is shown in Fig.~\ref{fig:ablation_2} the $L_{GAN}$ has an important rule in the enhancement of our proposed approach by transforming the polarimetric modality to the visible one. Moreover, adding facial attribute prediction loss enhances the face recognition performance. The reason behind this is because using facial attributes loss in addition to contrastive loss function leads to a more discriminative embedding space and this leads to a better face recognition performance. Consider a polarimetric subject with $Id \#2$ (see Fig.~\ref{fig3:latent}). The contrastive loss function causes the corresponding visible images from $Id\#2$ to move closer to $Id\#2's$ polarimetric and other $Ids'$ visible images to move farther away. Now, using the contrastive loss function in conjunction with the attribute classification loss function makes $Id \#1$ to move closer to $Id \#2$ since they share the same set of attributes (see Fig.~\ref{fig3:latent}). In other words, it differentiates between different impostors of $Id \#2$. The same procedure is performed for the other identities during the training process. Fig.~\ref{fig3:latent} visualizes the overall concept of adding facial attributes prediction loss function. As it is depicted, addition of attribute prediction loss leads to a more discriminative embedding subspace. This leads to a better face recognition performance as it is shown by the ROC curves in Fig.~\ref{fig:ablation_2}.    

\begin{table*}[]
\begin{center}
	\caption{Attribute prediction of the polarimetric face images using the proposed method and other frameworks and comparing it with the attribute prediction of visible faces.}
	\label{tableprediction}
	\scalebox{0.71}{
		\begin{tabular}{|c|c|c|c|c|c|c|c|c|c|c|}
			\hline
			Facial Attributes & Arched\_Eyebrows & Big\_Lips & Big\_Nose & Bushy\_Eyebrows & Bald & Mustache & Narrow\_Eyes & Beard & Mouth\_Slightly\_Open & Young \\ \hline
			Visible Input (Net A) &     96.7             &   98.4        &    99.1       &            95.9     & 99.3     &   99.4       &   95.7           &    98.9   &      97.7                 &  96.9  \\ \hline
			Polar Input (Net A) &  53.7                &  55.8         & 57.1          &     51.2            &  58.9    &   62.8       &   54.4           &  59.8     &      57.6                 &  52.7     \\ \hline
			Polar Input (fine-tuned A)&     78.2             &   83.9        &   85.3        &      80.7           &  88.4    & 89.3         &   79.3           &   88.9    &             81.9            &     76.5  \\ \hline
			Polar Input (Pol-GAN) &    89.6              &   95.3        &   96.6        &   90.4              &  96.2    & 95.2         & 91.9             &  94.8     &   93.7                    & 91.1       \\ \hline
		\end{tabular}}
	\end{center}
	
\end{table*}

\section {Attribute Prediction From Polarimetric Thermal}

One of the benefits of the proposed AGC-GAN is predicting facial attributes directly from polarimetric thermal modality. These attributes can be utilized directly or can be fused with other modalities to enhance recognition performance. In order to illustrate the effectiveness of the proposed method we performed attribute prediction in four different scenarios: 1) Attribute prediction of visible images with the VGG-Face based attribute predictor (Network $A$ in Sec.~\ref{attributess}). 2) Attribute prediction of polarimteric images with the attribute predictor $A$. 3) Attribute prediction of polarimetric images with the fine-tuned attribute predictor $A$. In this case, we fine-tuned the attribute predictor $A$ with the annotated polarimetric images and used it for the task of attribute prediction in the testing phase. 4) Attribute prediction of the polarimetric images using the Pol-GAN attribute predictor from the proposed AGC-GAN (see Fig.~\ref{fig2_ICB}). Table~\ref{tableprediction} shows the result of the prediction for the four mentioned frameworks. Although, fine-tuning the attribute predictor $A$ increased the prediction performance (framework $\#3$), but still its performance is less than our proposed framework. The proposed framework could outperform the other methods in predicting polarimetric face attributes and it has a comparable performance with the accuracy of predicted attributes from the visible face images (framework $\#1$). 

\section {Conclusion}
We have introduced a novel approach to exploit facial attributes information for the purpose of polarimetric thermal-to-visible face recognition. The AGC-GAN contains two GAN-based sub-networks dedicated to visible and polarimetric input images. The proposed network is capable of transforming the visible and polarimetric thermal modalities into a common discriminative embedding subspace and synthesizing the visible photos from the embedding subspace. We simultaneously minimize the cost functions due to the facial attribute identification in addition to other cost functions in order to increase the inter-personal variations between different subjects with different sets of facial attributes in the latent feature subspace. This leads to a more discriminative embedding subspace. An ablation study was performed to demonstrate the enhancement obtained by different losses in the proposed method. We compared our method with state-of-the-art polarimetric thermal-to-visible face recognition methods and showed the superiority of our proposed method over them.

{\small
\bibliographystyle{ieee}
\bibliography{test}
}

\end{document}